# Real-time Dynamic MRI Reconstruction using Stacked Denoising Autoencoder

Angshul Majumdar

Indraprastha Institute of Information Technology, New Delhi, India

## Abstract -

In this work we address the problem of real-time dynamic MRI reconstruction. There are a handful of studies on this topic; these techniques are either based on compressed sensing or employ Kalman Filtering. These techniques cannot achieve the reconstruction speed necessary for real-time reconstruction. In this work, we propose a new approach to MRI reconstruction. We learn a non-linear mapping from the unstructured aliased images to the corresponding clean images using a stacked denoising autoencoder (SDAE). The training for SDAE is slow, but the reconstruction is very fast - only requiring a few matrix vector multiplications. In this work, we have shown that using SDAE one can reconstruct the MRI frame faster than the data acquisition rate, thereby achieving real-time reconstruction. The quality of reconstruction is of the same order as a previous compressed sensing based online reconstruction technique.

## Introduction

In Magnetic Resonance Imaging (MRI) one pressing problem is to reduce the data acquisition time. In recent times Compressed Sensing (CS) based techniques are employed to achieve this end. The K-space is partially sampled (thereby reducing scan time); the image is recovered by exploiting its sparsity in some transform domain. CS based MRI reconstruction techniques have gained immense popularity in MRI reconstruction [1-17].

CS based reconstruction is slow, requiring solution of a non-smooth optimization problem that needs to be solved iteratively. This is not an issue for most scenarios including static scans (clinical diagnosis) or studies involving functional MRI. However there may be situations where we would like to recover the image on-the-fly, i.e. in real-time, e.g. in image-guided surgery or tracking problems (catheter tracking) [18, 19].

There are two approaches for causal real-time dynamic MRI reconstruction. The first approach is to apply Kalman Filtering techniques [18, 19] - these are fast, but do not yield very good quality images. The other approach is based on CS based techniques [20-23] - these improve reconstruction quality but are comparatively slow. CS based techniques cannot recover the images in real-time. The reconstruction methods proposed so far suffer from the obvious trade-off between the reconstruction quality and reconstruction speed.

If the K-space is uniformly under-sampled, and the image recovered by an Hermetian of the Fourier transform, we would notice aliasing artifacts. CS proposes randomly / incoherent under-sampling. If the

Hermetian of the Fourier transform is applied, we would still notice aliasing artifacts but these artifacts would be unstructured. It is not possible to design a linear inverse operator that can obtain an image from the partially sampled K-space - be it random or periodic undersampling. In every iteration CS does a non-linear thresholding operation that progressively removes the unstructured noiselike artifacts and extracts the image. Thus in essence CS is a non-linear inversion operator.

In this work we take a completely new approach to MRI reconstruction based on the Deep Learning framework - we will LEARN the non-linear inversion operation. More specifically our work is based on stacked denoising autoencoders (SDAE). During training the input to the SDAE will be the unstructured aliased (noisy) image (obtained by applying the Hermetian of the Fourier transform to the partially sampled K-space) while the output is the corresponding clean image. SDAE will learn the non-linear analysis (forward) and the synthesis (backward) mapping from the noisy image to the clean image. During testing the Hermetian of the Fourier transform is applied on the partially sampled K-space; the resulting noisy image is input to the SDAE and we will expect a clean version of it at the output.

Proponents of SDAE (and other deep learning tools like Restricted Botzman Machines and Convolutional Neural Networks) argue that, they can learn arbitrary non-linear inversion operations given enough data. The SDAE is not related to CS. But since CS is a non-linear inversion operator; we hope that SDAE will be a good tool to learn the non-linear inversion operation and yield good quality images.

SDAE requires a huge amount of data and very large training time (in days). However, once the non-linear mapping is learnt the run-time is very fast. It only requires a few matrix vector products (depending on the depth of the SDAE). Thus in practice, the learned SDAE can be used for real-time MRI reconstruction.

# Literature Review

## Dynamic MRI Reconstruction

Dynamic MRI reconstruction is an active area of research and there are a lot of papers on this subject. We will only cover some of the major techniques.

### Offline Reconstruction

The dynamic MRI data acquisition model can be expressed succinctly. Let $x_t$, denote the MR image frame at the $t^{th}$ instant. We assume that the images are of size N x N, but are vectorized; T is the total number of frames collected. Let $y_t$ be the k-space data for the $t^{th}$ frame. The problem is to recover all $X_t$'s (t=1…T) from the collected *k*-space data $y_t$'s. The MR imaging equation for each frame is as follows,

$$y_t = RFx_t + \eta \qquad (1)$$

where *R* is the sub-sapling mask, *F* is the Fourier mapping from the image space to the K-space, $y_t$ is the acquired K-space data, $x_t$ is the $t^{th}$ frame and η is white Gaussian noise.

A typical CS approach [1, 2] employs wavelet transform for sparsifying the spatial redundancies in the image domain and Fourier transform for sparsifying along the temporal direction. This is a realistic assumption since both [1, 2] of them worked on dynamic imaging of the heart where the change over time is quasi-periodic, thereby leading to a compact (sparse) support in the Fourier frequency domain.

The data from (1) can be organized in the following manner,

$$y = RFx + \eta \qquad (2)$$

where $y = \begin{bmatrix} y_1 \\ ... \\ y_T \end{bmatrix}$ and $x = \begin{bmatrix} x_1 \\ ... \\ x_T \end{bmatrix}$

In both [1, 2] the standard CS optimization problem was proposed to reconstruct the MR image data $x$ from the k-space samples $y$,

$$\min_x \|y - RFx\|_2^2 + \lambda \|W \otimes F_{1D}(x)\|_1 \qquad (3)$$

where $W \otimes F_{1D}$ is the Kronecker product of the wavelet transform (for sparsifying in space) and 1D Fourier transform (for sparsifying along temporal direction). In the original studies [1, 2] the Kronecker product was not used; we introduced it to make the notation more compact.

The 1D Fourier transform may not always be the ideal choice for sparsifying along the temporal direction (if the signal is not of periodic nature); in such a case 1D wavelet transform may be a better choice. Some other studies [3-5] assumed the image to be sparse in $x$-$f$ space, i.e. the signal was assumed to be sparse in $I \otimes F_{1D}$ basis (where $I$ is the Dirac basis). In these studies [3-5], the emphasis was on the solver for (3); they used the FOCUSS [6] method for minimizing (3). However, in practice any other state-of-the-art $l_1$-norm minimization solver (e.g. Spectral Projected Gradient L1 or Nesterov's Algorithm) will work.

Some other studies [7, 8] either assumed the MR image frames to be spatially sparse or were only interested in the 'change' between the successive frames. These studies did not explicitly exploit the redundancy of the MR frames in the spatial domain, rather they only applied a Total Variation (TV) regularization in the temporal domain. The optimization problem is the following,

$$\min_x TV_t(x) \text{ subject to } \|y - RFx\|_2^2 \leq \varepsilon \qquad (4)$$

where $TV_t = \sum_{i=1}^{N^2} \|\nabla_t x_i\|$ and $\nabla_t$ denotes the temporal differentiation for the $i^{th}$ pixel.

There is yet another class of methods that reconstruct the dynamic MRI sequence as a rank deficient matrix. The $k$-space acquisition model is the same as (1), but instead of concatenating the data as columns in (2), the data is stacked as columns in the following manner,

$$Y = RFX + \eta \qquad (5)$$

where $Y = [y_1 | ... | y_T]$, $X = [x_1 | ... | x_T]$ and $\eta = [\eta_1 | ... | \eta_T]$.

In [9, 10] it is argued that the matrix $X$ is rank deficient, since it can be modeled as a linear combination of very few temporal basis functions. Based on this assumption, [9, 10] proposed solving the inverse problem by matrix factorization,

$$\min_{U,V} \|Y - RF(UV)\|_F^2 \qquad (6)$$

where $X=UV$ such that $X$ is low-rank.

Solving the inverse problem via matrix factorization was an academic exercise. It did not yield the same level of accuracy as CS based techniques. However, later studies proposed combining the two [11-14]. In [12, 13] it was shown that good results are obtained when the rank deficiency of the signal in *x-t* space is combined with the sparsity in the *x-f* space. The following optimization problem was proposed to solve for the dynamic MRI sequence

$$\min_X \|Y - RFX\|_F^2 + \lambda_1 \|I \otimes F_{1D} \text{vec}(X)\|_1 + \lambda_2 \|X\|_* \tag{7}$$

In recent times, there are a few techniques that proposed learning the sparsifying basis in an adaptive fashion during signal reconstruction. This is dubbed as Blind Compressed Sensing (BCS) [15]. Here it is assumed that the dynamic MRI sequence can be sparsely represented in a learned basis by exploiting the temporal correlation, i.e. *X = ZD* where *D* is the sparsifying dictionary and *Z* is the coefficients. Unlike a fixed sparsifying basis like Fourier, *D* is learnt simultaneously with *Z*. The corresponding problem is formulated as:

$$\min_{D,Z} \|Y - F(ZD)\|_F^2 + \lambda_1 \|Z\|_1 + \lambda_2 \|D\|_F^2 \tag{8}$$

This formulation was proposed in [16]. An extension of this basic formulation was proposed in [17]. The first difference between [16] and [17] is that in the later, the sparsifying dictionary is learnt along the spatial direction and not along the temporal direction. Second, [17] exploits the low-rank structure of the dynamic MRI sequence ([16] handles it implicitly by controlling the number of atoms in the dictionary). Third, [17] proposes an analysis prior BCS formulation instead of the commonly used synthesis prior. The reconstruction was formulated via the minimization of the following problem [17]

$$\min_{D,X} \|Y - FX\|_2^2 + \lambda_1 \|DX\|_1 + \lambda_2 \|X\|_* + \lambda_3 \|D\|_F^2 \tag{9}$$

### Online Reconstruction

All the studies discussed so far are off-line reconstruction techniques. Online reconstruction techniques reconstruct the image corresponding to the current frame, given the reconstructed images till the previous time frame. In applications of dynamic MRI such as MRI guided surgery the reconstruction has to be performed in real-time. Online reconstruction does not guarantee a real-time solution. For being real-time, a reconstruction technique should be able to recover the images as fast as the frames are acquired.

Online dynamic MRI reconstruction is a new area of research and there are only a handful of papers on this subject. The most straightforward way to handle the online dynamic MR reconstruction is to use Kalman Filtering. That is exactly what is done in [16, 17] MRI reconstruction is formulated as a dynamical system and a Kalman filter model is proposed,

$$x_t = x_{t-1} + u_t \tag{10a}$$

$$y_t = RFx_t + \eta_t \tag{10b}$$

where the pixel values $x_t$ is the state variable, the *k*-space sample $y_t$ is the observation, $\eta_t$ is the observation noise and $u_t$ is the innovation in state variables.

In general the Kalman filter is computationally intensive since it requires explicit matrix inversion for computing the covariance matrix. However, in [18, 19], this problem was alleviated by diagonalization of the covariance matrix. This diagonalization was only possible under some simplifying assumptions. The problem with such a diagonal covariance is that the pixel values are assumed to change independently over time. This is not realistic. The diagonality assumption was also subject to sampling requirements – the sampling had to be non-uniform and non-Cartesian. In most practical systems, the sampling is Cartesian.

In [20] Kalman filtering of the wavelet coefficients is proposed. The filter model is the following,

$$\alpha_t = \alpha_{t-1} + u_t \tag{11a}$$

$$y_t = RFW^T \alpha_t + \eta_t \tag{11b}$$

where $\alpha_t$ is the wavelet transform coefficient vector for the $t^{th}$ frame.

This work was motivated by the findings of compressed sensing. The benefit of this approach is that wavelet whitens the spatial correlations, and hence the diagonality assumption is more feasible compared to the previous studies that operated in the pixel domain. The paper [20] proposed a further check over the Kalman update, i.e. when the error between the prediction and the actual data becomes large, instead of a filter update full reconstruction using CS technique is done.

In another work [21], it was proposed that instead of solving the image at the current time frame, the difference between the previous frame and the current one can be reconstructed instead. The following optimization problem is proposed,

$$\min_{\nabla \alpha_t} \| y_t - RFx_{t-1} - RFW \nabla \alpha_t \|_2^2 + \lambda \| \nabla \alpha_t \|_1 \tag{12}$$

where $x_{t-1}$ is the reconstructed image for the previous frame and $\nabla \alpha_t$ is the wavelet transform of the difference between the previous and the current frame.

The main assumption of [21] is that the sparsity pattern varies slowly with time. For each time frame, the difference of the wavelet transform coefficients $\nabla \alpha_t$ is computed, from which an intermediate estimate of the wavelet coefficients of current frame is obtained as,

$$\tilde{\alpha}_t \quad \nabla \alpha_t \tag{13}$$

The intermediate estimate is thresholded (via hard thresholding) and all the values below a certain threshold τ are discarded. The value of the thresholded wavelet coefficients are not of importance, but their indices are. Assuming that the set of indices having non-zero values after thresholding is Ω, the final value of the wavelet transform coefficient is estimated as,

$$\alpha_t = (RFW^T)^\dagger_\Omega y_t \tag{14}$$

where (•)$^\dagger$ denotes the Moore-Penrose pseudo-inverse and (•)$_\Omega$ indicates that only those columns of the matrix has been chosen that are indexed in Ω.

In [22] a simpler approach was proposed. It was postulated that the difference between subsequent frames is sparse. The sparse difference frame can be directly estimated using CS techniques and added to the previous frame.

$$\min_{\nabla x_t} \|y_t - RF(x_{t-1} + \nabla x_t)\|_2^2 + \lambda \|\nabla x_t\|_1 \qquad (15)$$

where $x_t = x_{t-1} + \nabla x_t$.

This simple technique yields good results in general. But this was improved further by replacing the previous frame as the reference frame by a linear prediction [23]. The assumption is that the difference between the linear prediction of the current frame and the current frame will be sparser than the difference between the previous frame and the current frame; such a difference could be recovered with fewer K-space measurements and hence improve temporal resolution of dynamic MRI.

## Autoencoder

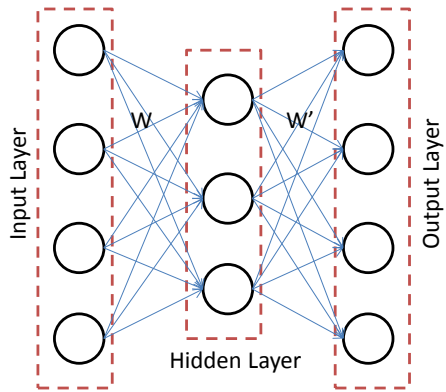

Fig. 1. Simple Single Layer Autoencoder

An autoencoder consists (as seen in Fig. 1) of two parts – the encoder maps the input to a latent space, and the decoder maps the latent representation to the data [24, 25]. For a given input vector (including the bias term) $x$, the latent space is expressed as:

$$h = Wx \qquad (16)$$

Here the rows of $W$ are the link weights from all the input nodes to the corresponding latent node. The mapping can be linear [26], but in most cases it is non-linear. Usually a sigmoid function is used, leading to:

$$h = \phi(Wx) \qquad (17)$$

The sigmoid function shrinks the input (from the real space) to values between 0 and 1. Other non-linear activation functions (like tanh) can be used as well.

The decoder portion reverse maps the latent variables to the data space.

$$x = W'\phi(Wx) \qquad (18)$$

Since the data space is assumed to be the space of real numbers, there is no sigmoidal function here.

During training the problem is to learn the encoding and decoding weights – $W$ and $W'$. In terms of signal processing lingo, $W$ is the analysis operator and $W'$ is the synthesis operator. These are learnt by minimizing the Euclidean cost:

$$\arg\min_{W,W'} \|X - W'\phi(WX)\|_F^2 \qquad (19)$$

Here $X = [x_1 | ... | x_N]$ consists all the training sampled stacked as columns. The problem (19) is clearly non-convex. It is solved by gradient descent techniques since the sigmoid function is smooth and continuously differentiable.

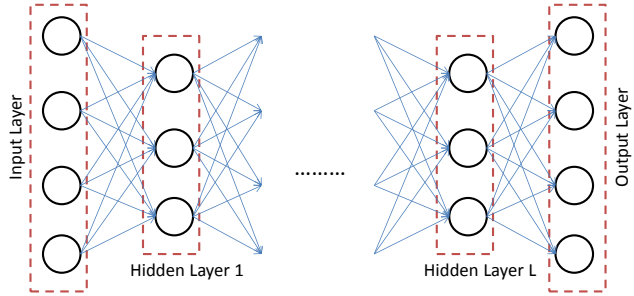

Fig. 2. Stacked Autoencoder

There are several extensions to the basic autoencoder architecture. Stacked / Deep autoencoders [27] have multiple hidden layers (see Fig. 2). The corresponding cost function is expressed as follows:

$$\arg\min_{W_1...W_{L-1}, W'_1...W'_L} \|X - g \circ \qquad (20)$$

where $g = W_1'\phi(W_2'...W_L'(f(X)))$ and $f = \phi(W_{L-1}\phi(W_{L-2}...\phi(W_1 X)))$

Solving the complete problem (20) is computationally challenging. The weights are usually learned in a greedy fashion – one layer at a time [28].

Stacked denoising autoencoders [29] are a variant of the basic autoencoder where the input consists of noisy samples and the output consists of clean samples. Here the encoder and decoder are learnt to denoise noisy input samples.

Another variation for the basic autoencoder is to regularize it, i.e.

$$\arg\min_{(W)s} \|X - g \circ \qquad + R(W, X) \qquad (21)$$

The regularization can be a simple Tikhonov regularization – however that is not used in practice. It can be a sparsity promoting term [30] or a weight decay term (Frobenius norm of the Jacobian) as used in the contractive autoencoder [31]. The regularization term is usually chosen so that they are differentiable and hence minimized using gradient descent techniques.

# Proposed Approach

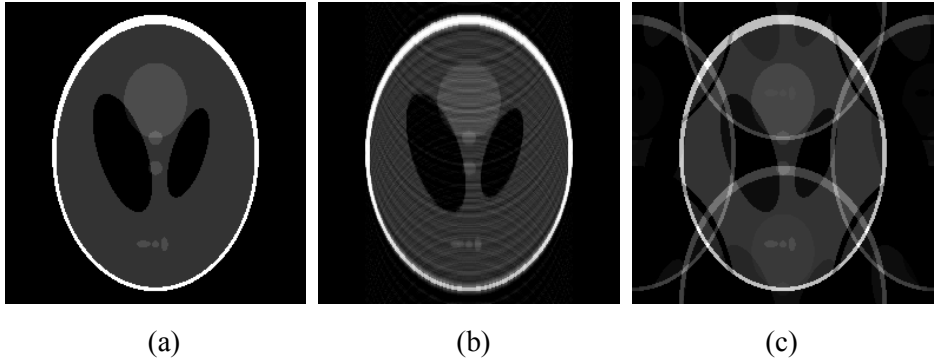

(a)            (b)            (c)

Fig. 1. (a) Original, (b)Reconstruction from Variable Density Random Sampling, (c) Reconstruction from Periodic Undersampling

In Fig. 1a. we show image of the Shepp-Logan phantom. Fig. 1b shows reconstruction by applying the Hermetian of the Fourier operator for variable density random sampling (a pattern popular in CS based MRI reconstruction). Fig. 1c. shows reconstruction by applying the Hermetian of the Fourier operator for uniform (regular) undersampling. We see that for regular undersampling, aliasing artifacts are present - this is expected. For variable density undersampling the aliasing is unstructured - it is more noiselike. Ideally, we would get even more unstructured artifacts if we used a completely random sampling mask, but this is not practically feasible for accelerating MRI scans.

The K-space data acquisition model is expressed as,

$$y = RFx + \eta \tag{22}$$

where the symbols have their usual meaning.

CS exploits the sparsity of the image in a transform domain for recovery. Assuming that the sparsifying transform is tight-frame or orthogonal, (22) can be expressed as,

$$y = RFW^T \alpha + \eta \tag{23}$$

Here $W$ is the sparsifying transform.

CS recovery algorithms based on $l_1$-norm minimization usually proceed in two steps -

- Gradient descent - which is a linear operation
- Thresholding - which is a non-linear operation.

These two steps are iterated till convergence.

Greedy algorithms for sparse recovery also proceed iteratively. The exact steps depend on the algorithm in use, but all of them have a non-linear operation either in the form of sorting (OMP [32]) or thresholding (Stagewise OMP [33]).

All CS recovery techniques are non-linear inversion operators. The main advantage of CS is that it is non-adaptive. But the disadvantage of CS is that it is iterative and hence time consuming; therefore not fit for problems where reconstruction speed is of concern. That is the reason, CS based techniques have not been very successful in real-time dynamic MRI reconstruction.

In this work, we propose to employ a stacked denoising autoencoder (SDAE) to learn the non-linear inversion in an adaptive fashion. As discussed before, autoencoders have a simple non-linear activation function associated with the neurons. Proponents of deep learning believe that autoencoders (and other related tools like Restricted Boltzman Machines) can learn arbitrary non-linear functional relationships given enough data. However there is no mathematical analysis to support this claim. In practice it has been found to give extremely good results. In some recent studies where SDAE was used for image denoising [26, 30] it was observed that SDAEs can compete with state-of-the-art denoising techniques like BM3D [34] and KSVD [35].

## Connection with Dictionary Learning

The KSVD has popularised dictionary learning in signal processing and machine learning. It has been used in almost every perceivable image processing scenario. The original KSVD formulation learns a dictionary that can represent the training data in a sparse fashion. The model is $X=D_S Z$ where $X$ is the data, $D_S$ is the learnt dictionary and $Z$ are the sparse coefficients.

In general dictionary learning can be expressed as,

$$\min_{D_S, Z} \|X - D_S Z\|_F^2 + \lambda \|Z\|_1 \tag{24}$$

The problem (24) is solved by alternately updating $D_S$ (dictionary / codebook) and $Z$ (sparse code).

This is the so called synthesis prior formulation where the task is to find a dictionary that will synthesize / generate signals from sparse coefficients. There is an alternate co-sparse analysis prior dictionary learning paradigm [36] where the goal is to learn a dictionary such that when it is applied on the data the resulting coefficient is sparse. The model is $D_A \hat{X} = Z$. The corresponding learning problem is framed as:

$$\min_{D_A, \hat{X}} \|X - \hat{X}\|_F^2 + \lambda \|D_A \hat{X}\|_1 \tag{25}$$

The autoencoder learns the analysis and the synthesis dictionaries simultaneously. The learning technique for a single layer autoencoder is (19); we repeat it for the sake of convenience.

$$\min_{W_A, W_S} \|X - W_S \phi(W_A X)\|_F^2 \tag{26}$$

It does not specifically look for a sparse solution (hence there is no sparsity promoting term). For the linear activation function, it is the same as jointly learning the analysis and the synthesis dictionaries.

## Design Principle

To design an autoencoder, there are two design aspects that needs to be specified - the number of hidden layers and the number of nodes in these hidden layers. In principle it is possible to learn arbitrary non-linear functional relationships with a single layer having a very large number of nodes - usually much larger than the dimensionality of the input samples. But this is not usually preferred. This is because, to learn a large number of weights the volume of training data needs to be very large. This is usually not available. Over the years, proponents of SDAE and other deep learning tools have proposed a stacked approach instead.

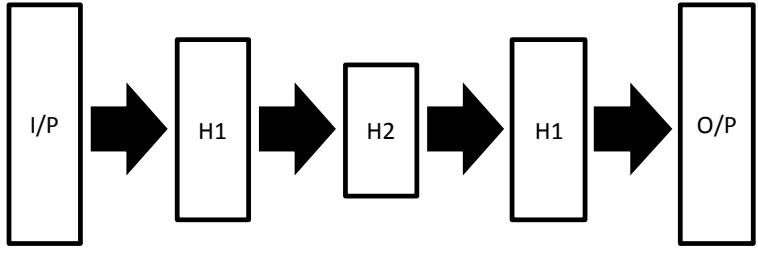

Fig. 2. Two Layer SDAE

Usually the number of nodes in the hidden layers reduce as one goes deeper into the SDAE. For example consider a document retrieval system; in [37] it was shown that starting from a 2000 length term frequency feature it is possible to learn a binary code (2 nodes) by using a multi-layer autoencoder with 500 neurons in the first layer, 250 in the 2nd hidden layer and 2 in the deepest layer.

To understand the learning process, we take a simple SDAE in Fig. 2. First the weights corresponding to the outermost hidden layer (H1) is learnt by solving the following problem.

$$\min_{W_A^1, W_S^1} \left\| X - W_S^1 \phi(W_A^1 X) \right\|_F^2 \tag{27}$$

This learns the weights and gives us the first level features $Z = \phi(W_A^1 X)$. The first level features are now used as training data for the second input layer. The weights for the second layer is solved by,

$$\min_{W_A^2, W_S^2} \left\| Z - W_S^2 \phi(W_A^2 Z) \right\|_F^2 \tag{28}$$

This process can be continued if there are more than 2 hidden layers.

There is no analytical rule that helps design an autoencoder. It is largely based on the experience of the researcher. Bengio and Hinton have independently suggested that care must be taken while reducing the number of nodes, the reduction should not be too fast or too soon.

## Proposed Architecture

In this work we are considering images of size 100 X 100. A variable density random sampling pattern is used for subsampling the K-space. The K-space data acquisition is modeled as (22). The input to the SDAE is the unstructured (noiselike) aliased image obtained by,

$$\hat{x} = F^H R^T y \tag{29}$$

The output consists of corresponding clean images. The SDAE learns the mapping from the noisy input to the clean output. During operation, we expect the SDAE to yield a clean image when presented with a noisy aliased image.

We propose to use a 2 hidden layer, 3 hidden layer and 4 hidden layer SDAE. The number of nodes in the first hidden layer is 2500, in the second hidden layer it is kept to 625 and in the third hidden layer it is kept to be 144. The sigmoid activation function is used throughout, i.e.

$$\phi(t) = \frac{1}{1 + e^{-t}} \tag{30}$$

It has been observed in [30] that Sparsity improves the denoising performance of autoencoders. Therefore we have learnt the weights of the deepest (bottleneck) layer using a sparsity promoting regularization.

$$\min_{W_A, W_S} \|X - W_S \phi(W_A X)\|_F^2 + \lambda \|W_A X\|_1 \qquad (31)$$

The issue with adding the sparsity promoting term is that we had to learn the SDAE for 2 layers, 3 layers and 4 layers separately.

# Experimental Results

The experimental data for training was obtained from the Laboratory of Neuro Imaging (LONI) at the University of Southern California. The dataset contains about 17424 volumes; and multiple slices in each volume. In total we have used about 100,000 images for training the SDAE's.

For testing we use the same datasets as in [22]. The experimental evaluation was performed on five sets of data. Two myocardial perfusion MRI datasets were obtained from [38]. These datasets were used in [8]. We will call these two sequences to be called the Cardiac Perfusion Sequence 1 and 2. The data was collected on a 3T Siemens scanner. Radial sampling trajectory was used; 24 radial sampling lines were acquired for each time frame. The full resolution of the dynamic MR images is 128 x 128. About 6.7 samples were collected per second. The scanner parameters for the radial acquisition were TR=2.5–3.0 msec, TE=1.1 msec, flip angle = 12° and slice thickness = 6 mm. The reconstructed pixel size varied between 1.8 mm2 and 2.5 mm2. Each image was acquired in a ~ 62-msec read-out, with radial field of view (FOV) ranging from 230 to 320 mm. Unfortunately, for these datasets, the fully sampled k-space scans are not available. Therefore the reconstruction results obtained from [8] is taken as the basis images for comparison.

The third and fourth datasets comprise of the Larynx and Cardiac sequence respectively. The data has been obtained from [39]. The larynx sequence is of size 256 x 256 and the cardiac sequence is of size 128 x 18 for each time frame. Six images were collected per second. We do not have further details pertaining to the MRI acquisition for these datasets. In this paper, follow the same protocol as in [22]; we have simulated radial sampling with 24 lines (for keeping it at par with the aforementioned dataset) for each image.

Our final dataset is obtained from [40]. It consists of a dynamic MRI scan of a person repeating the word 'elgar'. This is the Speech Sequence. The image is of resolution 180 x 180 and is obtained at the rate of 6 frames per second. This dataset was collected to study the tongue positions during speech. The MRI acquisition parameters related to this study are not reported here. For this sequence, we simulated radial k-space sampling with 24 lines.

For our experiments we have resized all the images to size 100 X 100 before simulating the K-space acquisition. This is done to maintain parity between our proposed SDAE and other techniques.

## Results on Reconstruction Accuracy

The reconstruction accuracy is measured in terms of Normalised Mean Squared Error (NMSE). The proposed technique is compared with one offline and two online techniques. The offline technique is k-t SLR [11]. The implementation for the same has been obtained from the authors' website [41]. The online techniques we compared against are Kalman Filtering [18] and Differential CS [22]. The author's of Kalman Filtering shared their code with us, we are grateful for them. The code of Differential CS is available with us. The reconstruction errors and the standard deviations are shown in the following Table.

Table 1. Mean and the Standard Deviation of the NMSEs on different datasets

| Dataset | k-t SLR | Kalman Filter | Diff CS | SDAE-2 Layer | SDAE-3 Layer | SDAE-4 Layer |
|---|---|---|---|---|---|---|
| **Cardiac Perfusion 1** | 0.123, ±0.040 | 0.340, ±0.112 | 0.179, ±0.067 | 0.207, ±0.098 | 0.183, ±0.056 | 0.181, ±0.045 |
| **Cardiac Perfusion 2** | 0.185, ±0.061 | 0.442, ±0.184 | 0.228, ±0.089 | 0.358, ±0.113 | 0.305, ±0.067 | 0.302, ±0.052 |
| **Larynx** | 0.028, ±0.012 | 0.048, ±0.020 | 0.032, ±0.015 | 0.043, ±0.021 | 0.038, ±0.015 | 0.035, ±0.010 |
| **Cardiac** | 0.075, ±0.022 | 0.127, ±0.101 | 0.079, ±0.028 | 0.0101, ±0.057 | 0.084, ±0.027 | 0.083, ±0.021 |
| **Speech** | 0.221, ±0.055 | 0.435, ±0.106 | 0.313, ±0.087 | 0.396, ±0.093 | 0.350, ±0.060 | 0.346, ±0.052 |

We observe that k-t SLR yields the best results. This is understandable - it is an offline technique and has access to the whole dataset for reconstruction. It exploits the spatio-temporal correlation from all the frames and therefore yields the best result. The results from Kalman filter are the worst. This has been observed before as well [22, 23]. The Kalman Filter makes certain simplifying assumptions (like motion being Gaussian and the pixels being uncorrelated) which simplifies the algorithm but in turn reduces the time accuracy. The Differential CS method yields good results, it is between k-t SLR and Kalman Filter. The problem with this approach is that the error keeps on accumulating; here the difference between every frame and the previous one is being reconstructed. The error incurred in reconstructing a frame is accumulated in subsequent frames therefore the error keeps on increasing with time.

Our proposed approach based on SDAE (2, 3 and 4 layers) yield higher error than Differential CS but is lower than the Kalman Filtering. We find that from the 2 (hidden) layer Sparse SDAE to 3 layer architecture, there is significant reduction in the mean and standard deviations of NMSEs. From the 3rd layer to the 4th layer there is only marginal reduction in mean NMSE but a significant improvement in standard deviation. The decrease in the standard deviation is owing to the averaging effect of the SDAE.

We have reported NMSE since it has been widely used as a metric for comparing MRI reconstruction. But NMSE or PSNR is not a good metric to represent subjective quality; SSIM is a better choice in that respect. In Table 2 we report the mean SSIM for all the datasets obtained from different reconstruction techniques.

Table 2. Mean SSIMs on different datasets

| Dataset | k-t SLR | Kalman Filter | Diff CS | SDAE-2 Layer | SDAE-3 Layer | SDAE-4 Layer |
|---|---|---|---|---|---|---|
| **Cardiac Perfusion 1** | 0.85 | 0.69 | 0.78 | 0.72 | 0.76 | 0.75 |
| **Cardiac Perfusion 2** | 0.87 | 0.66 | 0.80 | 0.73 | 0.79 | 0.77 |
| **Larynx** | 0.93 | 0.76 | 0.85 | 0.82 | 0.84 | 0.84 |

| Cardiac | 0.94 | 0.78 | 0.86 | 0.82 | 0.83 | 0.83 |
| Speech | 0.76 | 0.61 | 0.70 | 0.65 | 0.68 | 0.66 |

The results from Table 2 are almost similar to Table 1, except that we see the 4 Layer sparse SDAE actually yields slightly worse results than the 3 Layer structure. Although the 4 layer structure yields better results in terms NMSE, in terms of perceptual quality it is likely to be worse. For visual evaluation sample reconstructed images are shown in Fig. 1. The difference in reconstruction quality can be better evaluated from the difference images. These are shown in Fig. 2. The difference is magnified 10 times for visual evaluation.

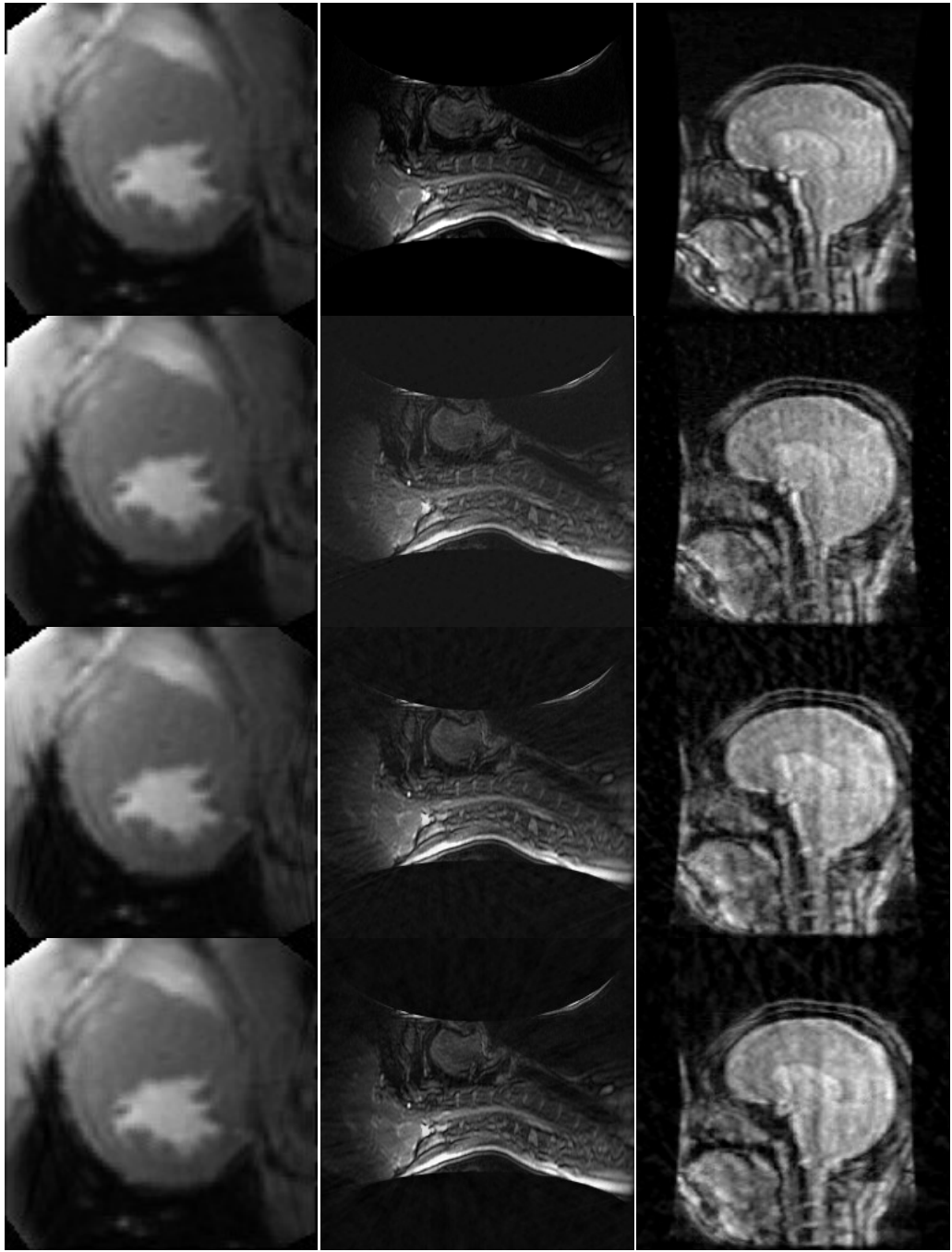

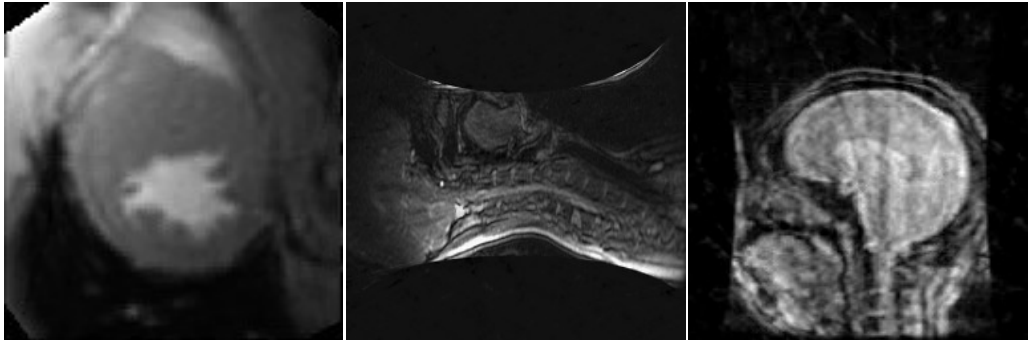

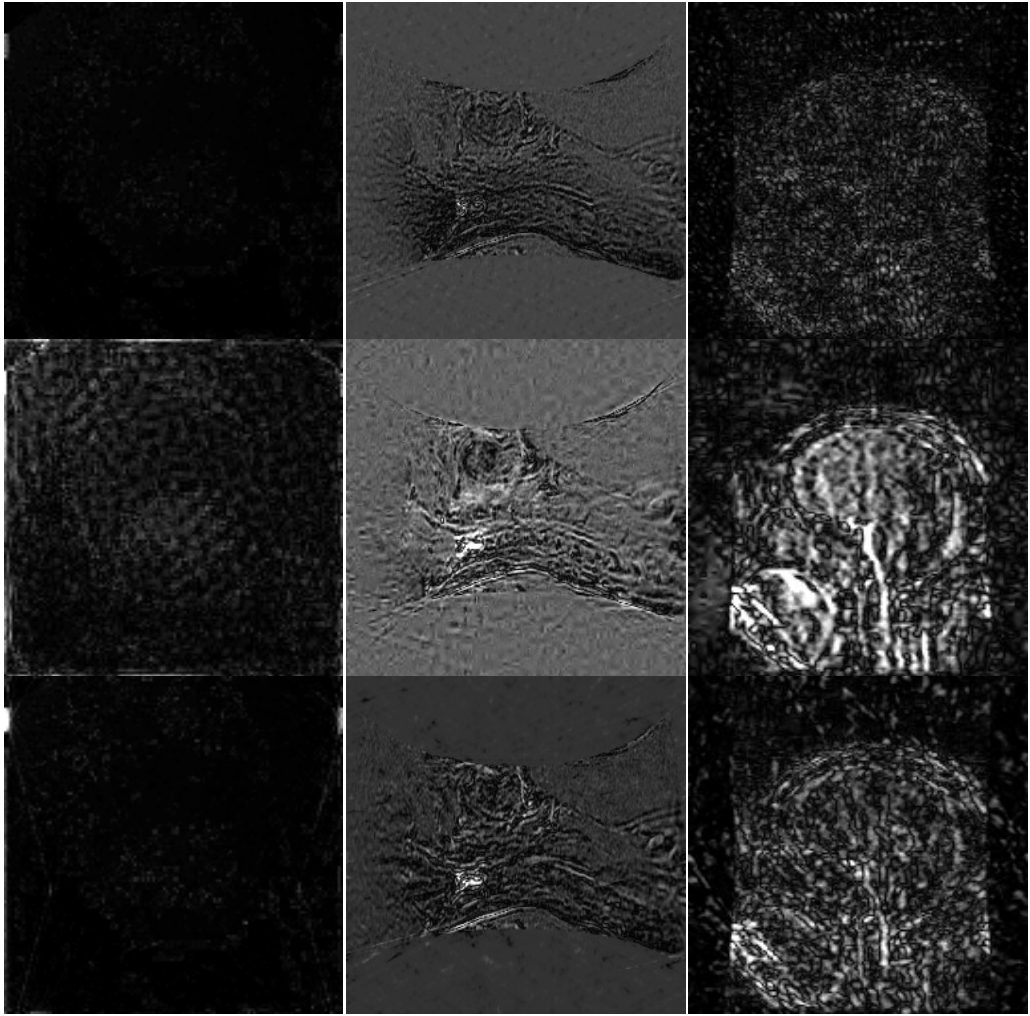

Fig. 1. Reconstructed Images. Left to Right - Cardiac, Larynx and Speech. Top to Bottom - Ground-truth; k-t SLR; Kalman Filter Reconstruction, Differential CS Reconstruction; Proposed SDAE Reconstruction (3 Layer)

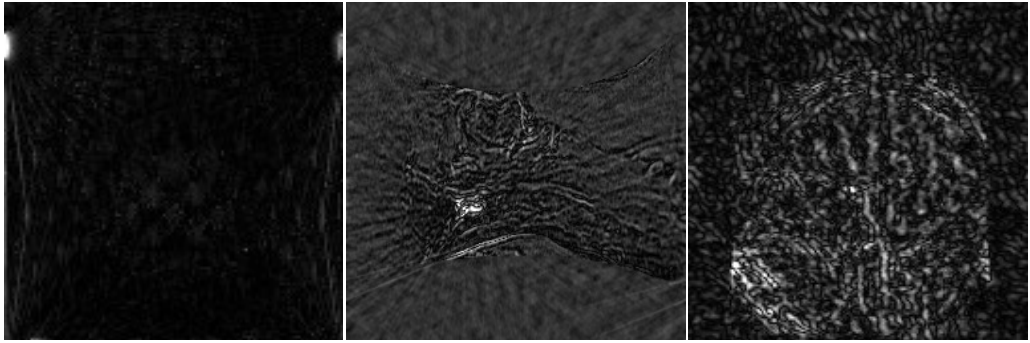

Fig. 2. Difference Images. Left to Right - Cardiac, Larynx and Speech. Top to Bottom - Ground-truth; k-t SLR; Kalman Filter Reconstruction, Differential CS Reconstruction; Proposed SDAE Reconstruction (3 Layer)

From the difference images it is clearly seen that k-t SLR yields the best results. Kalman Filter yields the worst reconstruction, there are a lot of reconstruction artifacts. The recovery results from the Differential CS formulation and our proposed SDAE are almost similar.

However we want to reiterate that Differential CS gradually accumulated error. Our method does not. We are independently reconstructing every frame. They are not part of any sequence. To emphasize it, we show the reconstructed and difference images of the speech sequence for the 3rd, 15th and 60th frames. One can see the error gradually increases with time for the Differential CS technique. We find that the error gradually increases for Differential CS - this is observed both from reconstructed and difference images. For our method (SDAE - 3 layer) the reconstruction is uniform.

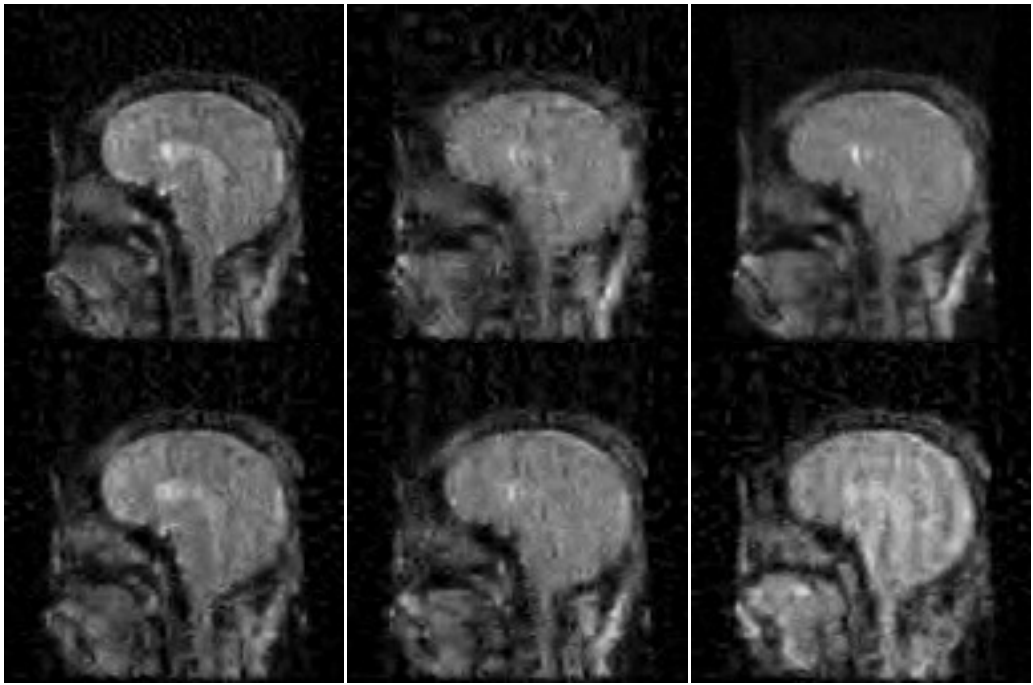

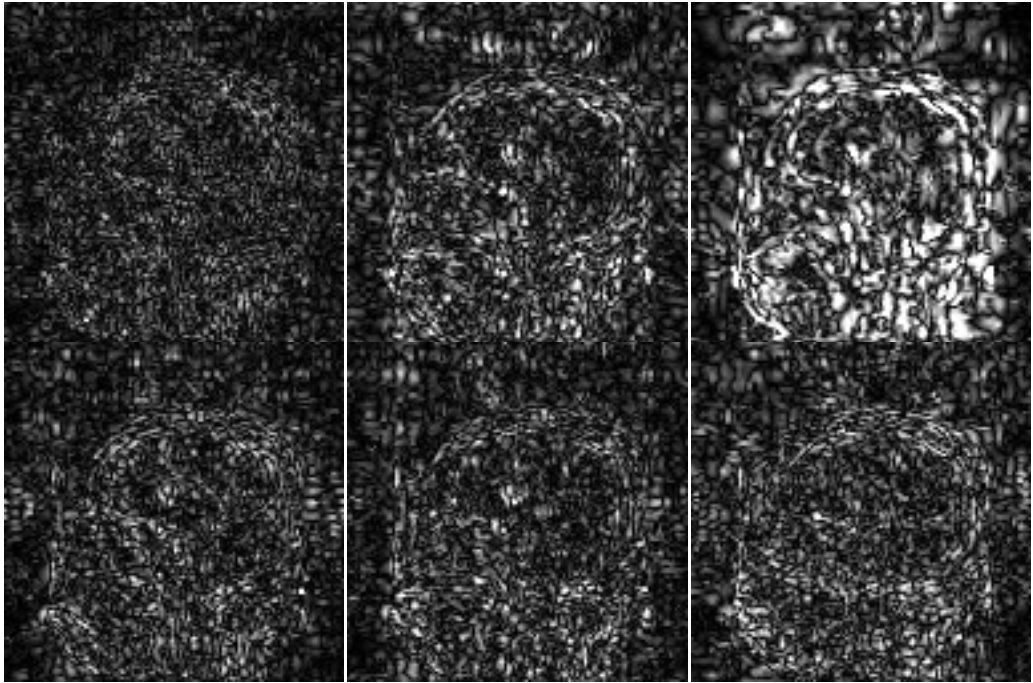

Fig. 3. Differential CS reconstruction and proposed reconstruction. Top row - reconstructed images for Differential CS; 2nd row - Reconstructed images from proposed method (SDAE - 3 layer). 3rd Row - difference images from Differential CS; Last Row - difference images from proposed reconstruction. Left to Right - 3rd, 15th and 60th frames.

## Results on Reconstruction Speed

Here we discuss the average reconstruction speeds from the three online techniques - Kalman Filter, Differential CS and Proposed. The experiments were run on a Core i7 CPU (3.1 GHz) having a 16 GB RAM running 64 Windows 7. The platform is Matlab 2012a.

As mentioned before, for all the experiments we have resized the images to size 100 X 100. The average reconstruction time from Kalman Filter is 0.33 seconds and from the Differential CS technique it is 0.13 seconds. For our proposed 2 layer SDAE the average reconstruction time is 0.0274 seconds, for 3 layer SDAE it is 0.0300 seconds 0.0310 seconds.

For images of size around 100 X 100 pixels, the data acquisition rate is usually between 6 to 7 frames per second (at full resolution). With acceleration (partial K-space sampling) this frame rate can go up higher - 18 to 21 frames per second for 3 fold acceleration. Studies like Differential CS can only reconstruct 7-8 frames per second. The Kalman Filter method is even slower (3 frames per second). Only our method can recover the frames at the required speed. SDAE only requires a few matrix vector multiplications (depending on the depth of the architecture); Our 3 layer SDAE can process 33 frames per second.

## Conclusion

This is the first work that proposes to frame MRI reconstruction from partial K-space samples in the stacked denoising autoencoder framework. Our technique requires a long training phase but a very short runtime. Hence, we believe it can be used for real time dynamic MRI reconstruction. We carried out extensive experiments to corroborate our proposal.

The major bottleneck of our approach is that it is only applicable to images of fixed size. We have assumed images to be size 100 X 100. The SDAE is learnt for this particular image dimension; to handle images of other sizes another SDAE needs to be learnt. However since training the SDAE is an offline process, it does not hamper the performance in any fashion.

Here we have assumed that the K-space sampling mask remains fixed and does not change with time. The autoencoder learns to get rid of the unstructured aliasing artifacts thus introduced. However, it is possible to change the sampling mask. In that case our proposed SDAE may not work and one may need to introduce training samples corresponding to all possible sub-sampling masks. In the future, we would like to work on this problem. We believe that learning an entirely new SDAE may not be necessary, and we can use domain adaptation concepts to handle the aliasing artifacts introduced by the variation in sampling masks.